\documentclass[runningheads]{llncs}
\usepackage[T1]{fontenc}
\usepackage{graphicx}
\usepackage{hyperref}
\usepackage{color}

\usepackage{amsmath}

\begin{document}
\title{Grid-Based Projection of Spatial Data into Knowledge Graphs}

\author{Amin Anjomshoaa\inst{1}\orcidID{0000-0001-6277-742X} \and
 Hannah Schuster\inst{1, 2}\orcidID{0000-0003-3032-1959} \and
 Axel Polleres\inst{1,2}\orcidID{0000-0001-5670-1146}
}

%
\authorrunning{A. Anjomshoaa et al.}

\institute{Vienna University of Economics and Business, 1020 Vienna, Austria \and
Complexity Science Hub Vienna, 1080 Vienna, Austria\\
\email{Amin.Anjomshoaa@wu.ac.at, Schuster@csh.ac.at, Axel.Polleres@wu.ac.at}}

\maketitle              
\begin{abstract}


The Spatial Knowledge Graphs (SKG) are experiencing growing adoption as a means to model real-world entities, proving especially invaluable in domains like crisis management and urban planning. Considering that RDF specifications offer limited support for effectively managing spatial information, it's common practice to include text-based serializations of geometrical features, such as polygons and lines, as string literals in knowledge graphs. Consequently, Spatial Knowledge Graphs (SKGs) often rely on geo-enabled RDF Stores capable of parsing, interpreting, and indexing such serializations. In this paper, we leverage grid cells as the foundational element of SKGs and demonstrate how efficiently the spatial characteristics of real-world entities and their attributes can be encoded within knowledge graphs.
Furthermore, we introduce a novel methodology for representing street networks in knowledge graphs, diverging from the conventional practice of individually capturing each street segment. Instead, our approach is based on tessellating the street network using grid cells and creating a simplified representation that could be utilized for various routing and navigation tasks, solely relying on RDF specifications. 


\end{abstract}
\section{Introduction}

Integrating geospatial data into knowledge graphs enriches the representation of real-world entities and fosters deeper insights into their spatial relationships. The resulting knowledge graphs, known as Spatial Knowledge Graph (SKG), are particularly useful in domains such as crisis management and urban planning, facilitating tasks such as proximity searches, resource planning, or spatial reasoning. Furthermore, SKGs enhance the understanding of relationships between entities, leveraging their geographic proximity or spatial interactions. The process of constructing SKGs from isolated spatial entities, such as landmarks, regions, points of interest, and observations, typically entails integrating their geometry serializations (i.e., as points or polygons) as well as their spatial relationships with other entities. 

While the Resource Description Framework (RDF) is currently used to capture, represent, and exchange the ever-growing volume of spatial and spatiotemporal data across diverse domains, RDF specifications offer limited support for managing spatial information effectively. Despite numerous efforts to represent and manage spatial data within RDF structures, as highlighted in \cite{zhang2021comprehensive}, these endeavors often resort to encoding geometrical features in text-based formats like Well-Known Text (WKT) or Geography Markup Language (GML). Consequently, the reliance on third-party implementations for spatial operations (e.g., topological relations such as neighborhood, incidence, and overlapping) and data processing has resulted in a lack of widely adopted and interoperable SKG solutions.

Beyond incorporating basic spatial features with geographic or spatial representations, typically depicted as points or polygons, integrating complex spatial networks like street networks into knowledge graphs is vital for specific applications, such as crisis management. However, despite the structural similarity between street network layouts and knowledge graph schema, encoding a projection of street networks that enables effective execution of network algorithms is not straightforward.

In this paper, we adopt a grid-based methodology to construct SKGs using basic RDF specifications tailored to meet the diverse demands of spatial applications across various scenarios and use cases. To this end, we leverage grid cells as the foundational element of SKG and demonstrate how efficiently the spatial characteristics of real-world entities and their attributes can be encoded within knowledge graphs. By following the Semantic Sensor Network (SSN) \cite{neuhaus2009semantic} and SOSA (Sensor, Observation, Sample, and Actuator) \cite{janowicz2019sosa} ontologies, this approach aligns with best practices for capturing measurements and observations at a fine-grained spatial and temporal dimensions. Another contribution of this paper is the introduction of a novel methodology for representing street networks in knowledge graphs, diverging from the conventional practice of individually capturing each street segment. Instead, our approach is based on tessellating the street network using grid cells and creating a simplified representation that could be utilized for various routing and navigation tasks.

\section{Methodology}

\begin{figure}
\includegraphics[width=\textwidth]{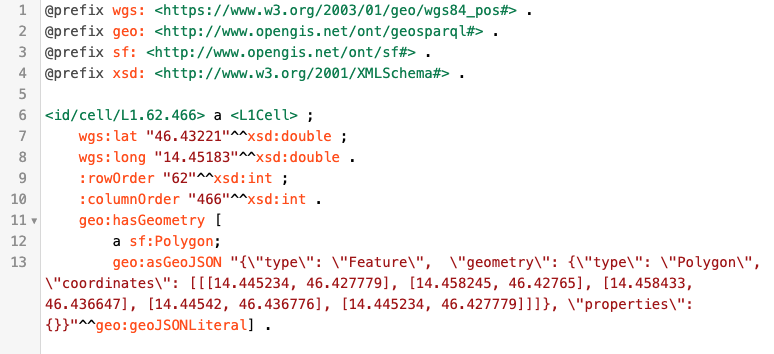}
\caption{Definition of grid cell <id/cell/L1.62.466>.}  \label{cell_definition}
\end{figure}

We utilize a grid cell system to organize spatial features and their relationships. Once organized into grid cells, relevant attributes will be expanded and inferred for the overlapping real-world geometries, such as communities, cities, and states. The selected grid cells in our work are one square kilometer cells defined by MGI/Austria Lambert(EPSG:31287) projection, a spatial reference system for Austria. In this context, every grid cell includes designated row and column orders, along with the precise coordinates of its center. The row and column orders facilitate the process of querying grid cells within a specified range and simplify the retrieval of relevant grid cells. Additionally, each grid cell is associated with a polygon geometry used for visualization purposes. Figure \ref{cell_definition} shows the definition of a gird cell <id/cell/L1.62.466> with specific row and column orders.

We leverage these cells as the fundamental building blocks of our SKG in order to effectively manage diverse spatial operations as described below:  

\begin{itemize}
\item The grid cells can be used to construct larger regions, including political regions (e.g. communities, districts, states) and service regions (e.g. hospital and fire-brigade care zones). In these cases, the regions are defined as spatial features and mapped to underlying grid cells using existing spatial relationships (e.g., using sfWithin property defined in GeoSPARQL Ontology). 

\item Spatial entities, including landmarks, facilities, and points of interest, are directly mapped onto grid cells. Consequently, depending on particular use cases, we can infer relationships between these entities and overlapping regions by querying the corresponding grid cells of each region. Moreover, entities assigned to grid cells may span multiple overlaying regions, and their allocation remains unaffected by changes in the definition of regions, which might change over time (e.g., changes in care regions of hospitals or fire brigade stations). This ensures the robustness and flexibility of our spatial analysis, allowing us to maintain consistency and relevance regardless of geopolitical adjustments.

\item Grid cells also include concepts and properties describing observations and relevant spatiotemporal data, like weather and climate data, social media, and demographic information. To this end, we can follow the best practices in semantic sensor networks and lightweight observations modeling techniques \cite{janowicz2019sosa} to include such data. For instance, Figure \ref{cell_observation} depicts a spatiotemporal observation that defines a specific grid cell's measured global radiation value. 

\begin{figure}
\includegraphics[width=\textwidth]{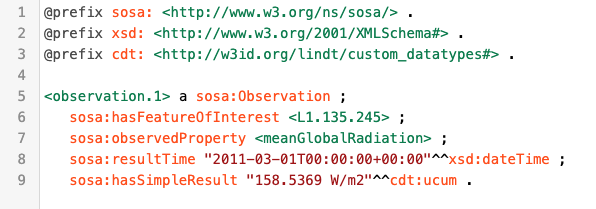}
\caption{Representing cells observation following SOSA ontology.}  \label{cell_observation}
\end{figure}

\item Finally, grid cells can include a range of indicators pertaining to their associated infrastructure networks, including street networks, railways, water systems, and power networks. To achieve this goal, grid cells partition the overlapping networks into smaller segments. Then, the infrastructure networks can be defined based on interconnected relationships and dependencies between such segments. To this end, through systematic processing, pertinent indicators and critical data are extracted and integrated into the definition of each grid cell. These enriched definitions empower spatial processes and operations by providing comprehensive insights into the infrastructure networks.
\end{itemize}

In the rest of this section, we first introduce a specialized grid cell indicator, namely Cell Orientation, constructed based on the properties of the underlying street network. Finally, we utilize the introduced grid cell indicators and spatial scaling operations to realize routing and navigation use cases, demonstrating how grid cells can contribute to creating a simplified street network that facilitates such scenarios.



\subsection{Grid Cell Orientation}

We define the orientation of grid cells as a quantitative indicator representing the efficiency of traffic flow within them. This indicator relies on various properties of the underlying street network, including the direction and number of roads in different orientations. This orientation indicator of cell $\mathcal{C}$ in different directions is calculated based on the overlapping street segments ${S_i}$ with cell $\mathcal{C}$ as follows: 

\begin{equation}
    \mathcal{O}_d(C)=\begin{cases}
    \sum_{S_i \in C} {L_{S_i} * \cos(B_{S_i}) * W_{S_i}}, & d = \text{East or West} \\
    \sum_{S_i \in C} {L_{S_i} * \sin(B_{S_i}) * W_{S_i}}, & d = \text{North or South}
    \end{cases}
\end{equation}

Where $L_{S_i}$, $B_{S_i}$, $W_{S_i}$ are length, bearing angle, and number of street lanes for each street segment, respectively and all these values are readily available in Open Street Map data. The orientation of a street network plays a pivotal role in shaping traffic flow within a city or urban area. For instance, grid-based layouts with perpendicular streets promote a balanced and straightforward traffic distribution, enabling smoother navigation. On the contrary, the street networks in older cities often direct traffic towards a central focal point, typically centered around historical landmarks like churches or communal gathering spaces. While the orientation and entropy of street networks have been extensively studied in the context of cities \cite{boeing2019urban}, there is a noticeable gap in the exploration of grid-based orientations and their applications. Existing research has predominantly focused on the entire street network of target cities, which basically overlooks the unique properties of grid cells and their interplay with adjacent cells. The orientation indicator relies on several factors of the underlying street network, including the orientation, width, and number of roads in various directions.

In order to build orientation and connectivity of grid cells, we use Open Street Map data and extract the major road types, including motorway, trunk, primary, and secondary road segments for each grid cell. The rationale behind selecting these road types lies in their pivotal role as the primary transportation backbone. To this end, we extract the street network corresponding to each grid cell and calculate the effectiveness of road segments towards north, south, east, and west directions. For the directional effectiveness of each road segment, we use the bearing of road segments and assess the alignment of the road vector with respect to the latitude and longitude axis directions. Moreover, we incorporate a weight for each street segment to denote its transportation capacity, factoring in the number of lanes and the maximum speed limit. We refer to this indicator as Network Orientation Indicator and denote it as $\mathcal{O}$.

Please note that the $\mathcal{O}$ indicator is upward scalable, which means to assess the $\mathcal{O}$ of larger regions $R$ in direction $\theta$ (i.e., north, south, east, and west), we can aggregate the $\mathcal{O}$ values of smaller cells in that direction, which could be calculated as shown in the following equation: 

\begin{equation}
    \mathcal{O}_\theta(R) = \sum_{i=1}^n \mathcal{O}_\theta(R_i).
    \label{noi_aggregation}
\end{equation}

Finally, the grid cells in our knowledge graph will be enriched with measurable orientation metrics for evaluating the effectiveness of traffic dynamics across grid cells. This multifaceted application enhances our understanding of both spatial relationships of grid cells within the knowledge graph and the practical implications of traffic patterns in the real-world context.

\subsection{Routing and Navigation}

To facilitate routing and navigation functionalities, we construct a simplified representation of the street network within each grid cell. This representation encompasses a compact yet comprehensive view of the grid cell's function within the larger road network, including routing and traffic flow dynamics. To this end, we leverage the Open Street Map (OSM) data again and specifically extract major road classifications, including motorways, trunk roads, primary routes, and secondary roads within each grid cell. Notably, this selection is not limited to these primary categories; it can be expanded to include smaller road classifications and diverse forms of mobility networks, including roadways and water routes.

In order to construct the simplified network, we first define terminal edge and terminal node concepts. Let $\mathcal{N}$ be the complete street network of a target region. Then, the portion of the street network that overlaps with grid cell $C_i$ is denoted by $\mathcal{N}|_{C_i}$ and includes all nodes and edges of the street network that are located inside the boundaries of grid cell $C_i$. The terminal edges of grid cell $C_i$, denoted as $\mathcal{T}^{\to}_{C_i}$, consist of the street segments in network $\mathcal{N}$ where the start node belongs to the grid cell network $\mathcal{N}|_{C_i}$ and the end node does not: 

\begin{equation}
\mathcal{T}^{\to}_{C_i} = \{ e_i = (u, v) \mid (u \in \mathcal{N}|_{C_i} \land v \notin \mathcal{N}|_{C_i})).
\end{equation}

In other words, the terminal edges include the street segments that we use to cross the borders of grid cell $C_i$ and enter the neighboring cells. It is clear that for a two-way street segment, we will have two terminal edges, one for outgoing traffic and one for incoming traffic.

Similarly, we define terminal nodes of grid cell $C_i$ to be the last nodes visited inside $\mathcal{N}|_{C_i}$ before leaving the grid cell network via one of its terminal edges and denote it as $\mathcal{T}^{\circ}_{C_i}$. So, according to this definition, every terminal node of grid cell $C_i$ is connected to a terminal node of a neighboring cell via a terminal edge.

\begin{figure}
\includegraphics[width=\textwidth]{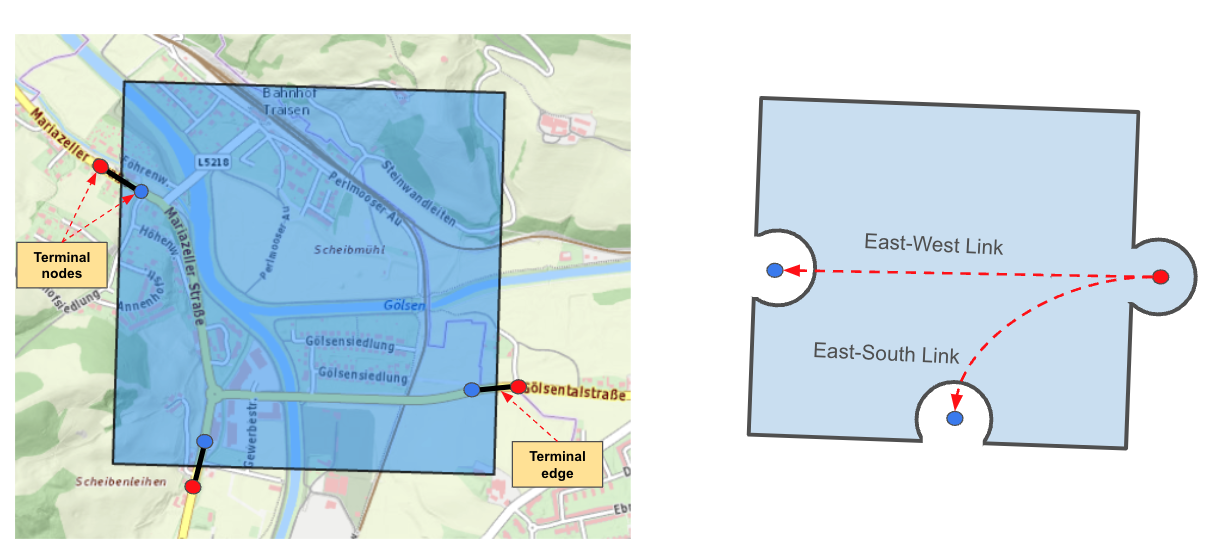}
\caption{Representing cells as puzzle pieces with links.}  \label{puzzle}
\end{figure}

Now, we use the terminal nodes to replace the grid cell network $\mathcal{N}|_{C_i}$ with a minimal yet functional street network that connects the neighboring cells via terminal nodes. This minimal street network can be metaphorically represented as a puzzle piece, where the terminal nodes serve as connectors on each side. Figure \ref{puzzle} illustrates a sample grid cell along with its terminal edges and nodes, alongside the resulting puzzle piece constructed from these terminal nodes.

In order to create the minimal street network of each cell, during the construction phase of KG, we analyze the corresponding grid cell network $\mathcal{N}|_{C_i}$ and examine the feasibility of routing between each terminal node of target cell paired with the terminal nodes of neighboring cells. If there is a route between this pair of nodes, we create a link between those terminal nodes.

\begin{figure}
\includegraphics[width=\textwidth]{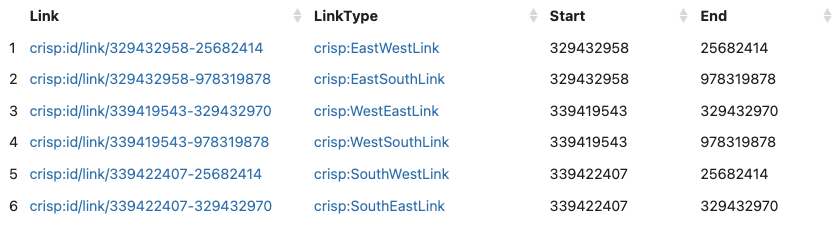}
\caption{Sample links generated for a specific cell.}  \label{links}
\end{figure}

During the construction phase of KG, we establish the minimal street network of each cell based on the respective grid cell network $\mathcal{N}|_{C_i}$ and its terminal nodes. To this end, we assess the feasibility of routing between each terminal node of the target cell and the terminal nodes of neighboring cells. If a route exists between such a pair of nodes, we establish a link between them and store the link in KG accordingly. These links also incorporate a property indicating the direction of the link (Link Type), which is essential for facilitating routing and navigation within the target cell. Figure \ref{links} shows the constructed links for the grid cell presented in Figure \ref{puzzle}. Please note that for two-way connections between terminal nodes, we would require two links in opposite directions to facilitate bidirectional connectivity.

\begin{figure}
\includegraphics[width=\textwidth]{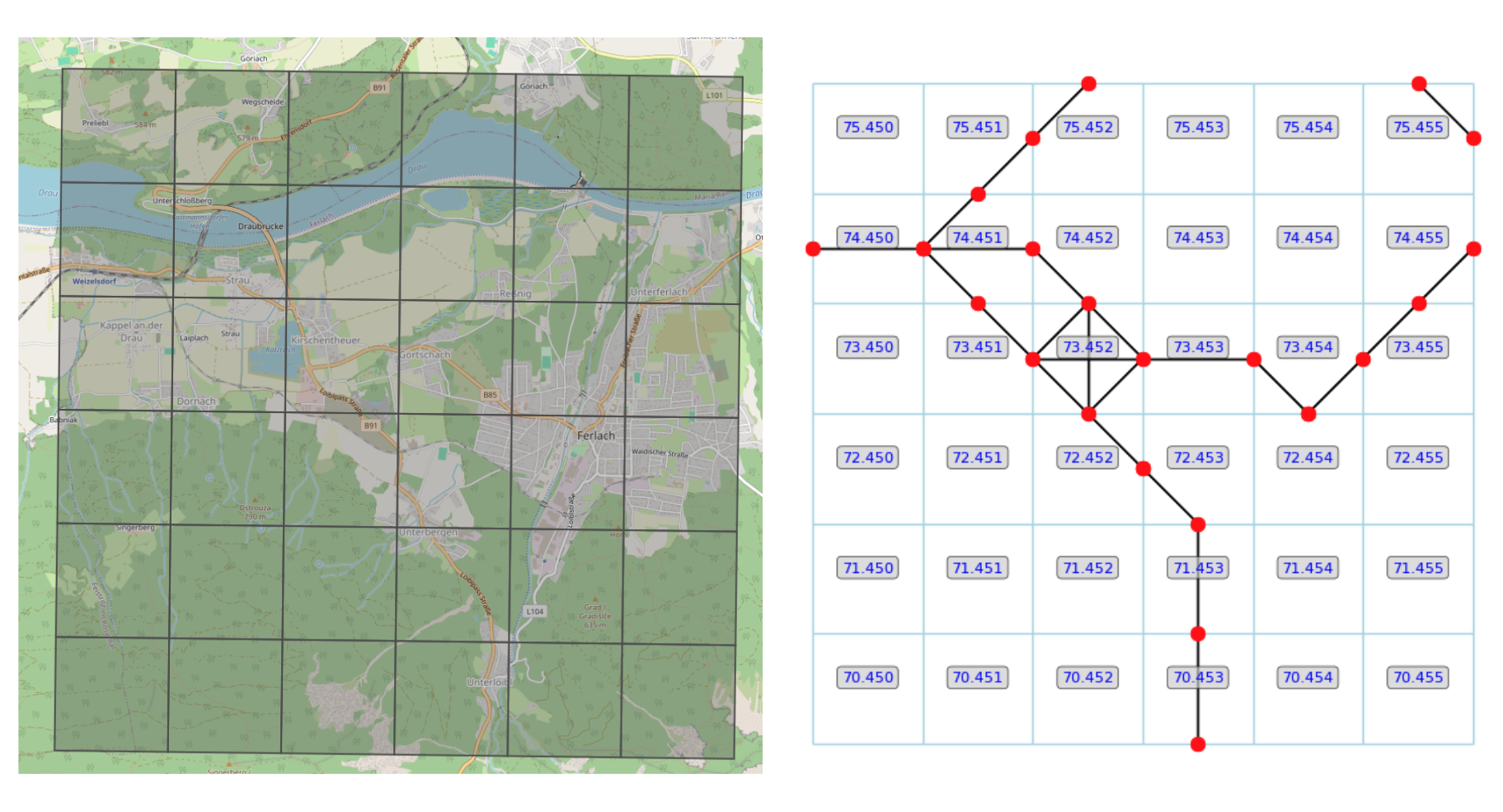}
\caption{Mapping grid cells to simplified cell network for navigation use case.}  
\label{grid_mapping}
\end{figure}

Now, in order to navigate between each pair of cells, we simply connect their corresponding puzzle pieces together, constructing a simplified street network. Figure \ref{grid_mapping} shows several grid cells alongside their corresponding simplified street networks generated using our methodology.

The simplified street network is suitable for employing established graph routing and navigation algorithms akin to those used in traditional routing on larger street networks. However, a key distinction lies in the efficiency enabled by our minimal street network facilitated through SKG. We can seamlessly retrieve the links via straightforward SPARQL queries and swiftly construct the minimal street network on the fly. 

For instance, let's consider applying the $A^*$ algorithm \cite{hart1968formal}, a widely-used graph traversal and pathfinding algorithm, to navigate through the minimal street network depicted in Figure \ref{grid_mapping}. 

During the iterations of the $A^*$ algorithm in the context of our simplified street network, we need to determine the next grid cell to extend. $A^*$ does this based on the cost of the path from the start node to the current node ($G$) as well as the cost from the current node to the target node ($H$) and then selects the path that minimizes the total cost ($F = G + H$). In the context of our SKG, we can compute the traversal cost between each pair of cells as the distance between those cells, which can be naively computed based on their row and column orders. However, we've already integrated a more efficient indicator into our SKG: the network orientation indicator $\mathcal{O}$. This indicator can be utilized to estimate the traversal cost between each pair of cells, and as discussed before, it is upward scalable. Depending on the location of the target cell (or origin cell) relative to the current cell, we can aggregate the relevant orientation indicators. This aggregated value can then be considered a more realistic cost for moving towards the target cell. For instance, if the origin cell is situated southeast of the current location, we begin by querying the cells between the origin cell and the current cell based on row and column order ranges. Then, we consider the sum of those cells' west and north orientation indicators as the traversal cost ($G$) from the origin cell. This step will be repeated for the current cell and the target cell as well, enabling us to calculate the traversal cost to the target ($H$). Finally, we compute the total traversal cost ($F$) by summing up $G$ and $H$.

\section{Grid Knowledge Graph Applications}

\begin{figure}
\includegraphics[width=\textwidth]{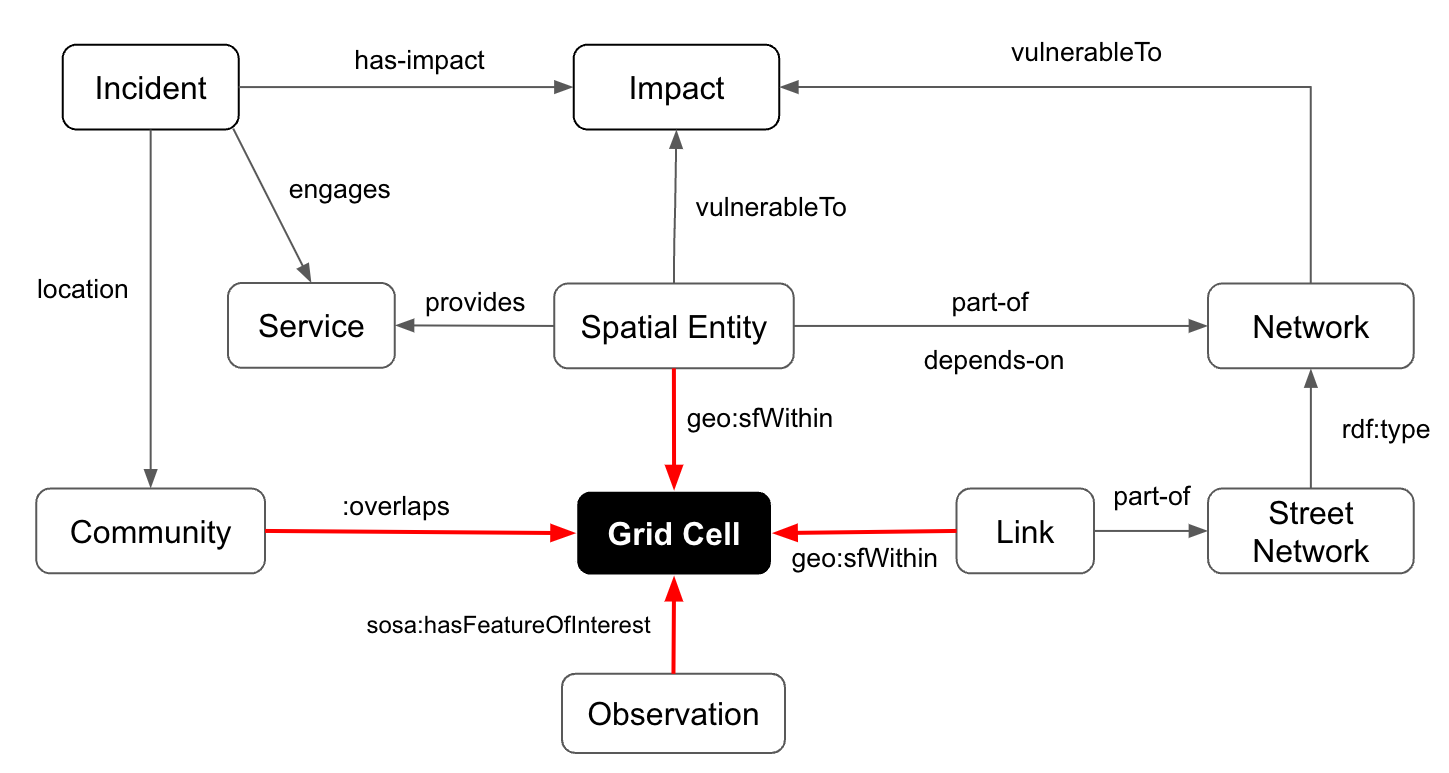}
\caption{The semantic model of Spatial Knowledge Graph for crisis management use case, highlighting the geo-related properties that facilitate spatial operations.}  
\label{crisp_data_model}
\end{figure}

To illustrate the application of our approach, we applied the proposed grid system in the context of CRISP crisis management project \footnote{\url{http://crisp.ai.wu.ac.at/}}, which aims to become a national Information Integration Hub for Crisis Management. The project's SKG is already available as an online resource, complete with a dedicated website and SPARQL endpoint. In the context of this project, we're addressing a critical issue in crisis response and intervention by enhancing access to vital data during emergencies. Accessing uniform and concise data sources poses one of the most significant challenges during crisis situations. Valuable data is often fragmented and scattered across various systems, organizations, and agencies, making it challenging to obtain a comprehensive understanding of the situation. However, besides data accessibility, crisis response is often hindered by the complexity and interconnectivity of our built environment as it draws upon various infrastructure networks, such as the communication network, the power network, and the transportation network. Consequently, understanding the interdependencies within geospatial networks is crucial for effective crisis management as it provides valuable insights into the dynamics of such situations. Especially considering that if an element of one system fails, it may spread issues within that system or cause issues in others, sometimes leading to cascading effects. Figure \ref{crisp_data_model} depicts the data model of our crisis management KG and highlights the central role of grid cells in data integration and spatial operations.

 So far, analyzing the resilience of a street network and its behavior during crisis events primarily focuses on weaknesses and bottlenecks using network analysis tools. Even though these results can be used to improve response planning, they only look at one aspect of the problem, and the application of network analysis is not viable in the midst of a crisis. In this section, we demonstrate the utility of our novel methodology for representing street networks in a KG. Given that the state of road networks directly and significantly influences crisis management and the subsequent outcomes of a crisis event, our approach is geared towards enabling swift analysis of these complex networks to identify vulnerable grid cells, allowing for quick assessments on the go. Figure \ref{crisp_data_model} depicts the semantic model of our SKG for crisis management use case. It shows the integral role of grid cells for connecting spatial entities (depicted with red arrows) ranging from observations and points of interest (such as hospitals and fire stations) to regions and road networks.

 Furthermore, given the complexity of the road network, a simplification becomes necessary, particularly when conducting a comprehensive analysis of the road network of an entire country. Different approaches exist to accomplish this, for example, by using municipalities as nodes and aggregating the connecting streets into single links. By using grid cells for simplification, we have the capability to aggregate data into larger blocks, simplifying the road network further while simultaneously considering the grid cells as the latest fundamental unit to analyze areas of interest in more detail.

 Besides identifying possible at-risk grid cells, this approach also enables routing during crisis events. The accessibility of the affected region for the different first responders, as well as the accessibility of critical infrastructure from the impacted zone, are additional use cases covered by our approach. By eliminating disaster-impacted cells from the KG, we are able to investigate alternative routing and navigation possibilities for the street network. Moreover, using the grid cells enables an easy extraction of the impacted area along with a buffer zone, allowing for a focused analysis of the involved infrastructure. This permits a more detailed and in-depth analysis of the road network and services depending on it.

\section{Discussion}

The grid-based methodology proposed in this research opens avenues for constructing SKGs that operate independently of the necessity for geo-enabled RDF stores. It has demonstrated the potential to meet the spatial application requirements across diverse use cases, including crisis management. Furthermore, grid cells offer numerous advantages for capturing spatial entities and scaling quantitative indicators. Despite the time-consuming nature of constructing simplified street networks for grid cells, it's a one-time computation. Once generated, the network and calculated cell properties remain largely unchanged, enabling long-term usage for routing and navigation applications.

Another topic requiring further exploration is the choice of grid cell shape and size. As previously mentioned, we utilize square-kilometer grid cells in our solution. This decision was driven by the availability of national weather data at this granularity level. However, we've found that this grid size is also beneficial for other crisis management applications, such as capturing infrastructure networks and spatial entities. Nevertheless, the proposed SKG has the capability to support multiple grid cell sizes simultaneously. For example, we could incorporate larger grid cells, such as 100 square kilometers, built on top of the smaller grid cells. These larger cells may also support upscaling and downscaling operations, as discussed in this research. Regarding the shape of the grid cells, an alternative approach could involve using other space-filling polygons, such as hexagons, instead of the currently used rectangles. In fact, hexagonal grid cells have already been explored for various urban computing applications and have shown advantages, particularly in navigation use cases. Uber's Hexagonal Hierarchical Spatial Index (H3) is one such solution that appears promising for navigation and geospatial analysis of street networks. However, it's worth noting that since hexagonal grid cells cannot be fully covered by smaller hexagons, scaling quantitative indicators would not be a straightforward task.

During the computation of cell connections, we encountered several issues that needed to be addressed differently. One such issue involved terminal nodes located exactly on the border between grid cells. To resolve this, we simply removed such points and introduced the immediate nodes on the street segment as the terminal nodes. Another issue arises with complex street segments that cross the cell border multiple times before continuing into neighboring cells. Such occurrences are more prevalent in mountainous terrain. To resolve this issue, we need to adjust the terminal nodes again to eliminate the irregularities in the street segments.

Another unaddressed issue in this research is the challenge of last-mile routing and navigation. Our proposed solution offers a high-level routing approach and does not encompass detailed step-by-step navigation instructions. This limitation stems from two main reasons. Firstly, we filtered out smaller roads during the construction of grid cell networks. Secondly, we simplified the street network to prioritize connections between grid cells exclusively. Regarding the first issue, we could address it by including all available road types and investing more time and computing power to compute the grid cell networks. However, it's important to note that we intentionally avoided working with the complete street network, which offers several advantages for our target domain, namely crisis management. Nonetheless, for other applications, detailed routing and navigation may still be necessary. A potential workaround for this problem could involve storing a serialization of each cell's street network in the SKG. This solution is akin to existing approaches that incorporate geometrical features in the KG using text-based formats like Well-Known Text (WKT) or Geography Markup Language (GML). Consequently, whenever necessary, we can retrieve the detailed street network of each grid cell and provide detailed navigation instructions.

\section{Related Work}

One of the earliest efforts to integrate geospatial information into RDF is the W3C Basic Geo Vocabulary \cite{w3cgeo}. It offers the Semantic Web community a namespace for representing spatial features represented by points, with WGS84 serving as the reference datum. This was later followed by GeoRDF proposal, aiming to offer an RDF-compatible profile for describing geometric information, including points, lines, and polygons. Over the past decade, the Semantic Web Community has continued this line and joined forces with the Open Geospatial Consortium (OGC) to make spatial data more effectively available and address requirements in the representation, query, and storage of spatial and spatiotemporal RDF data. A noteworthy result of this joined effort is the GeoSPARQL ontology \cite{geosparql}, which defines an RDF/OWL vocabulary for representing spatial information. This ontology has been widely used for the representation and querying of geospatial linked data and supported by a number of RDF store implementations, including prominent ones like Apache Jena, Ontop VKG \cite{bereta2019ontop}\cite{xiao2020virtual}, and Strabon \cite{kyzirakos2012strabon}, to mention just a few. Many of these implementations rely on the spatial functionality and spatial indexing offered by geo-enabled databases like PostGIS. While these databases are generally efficient, they can encounter challenges with specific operations, such as GeoSPARQL filter queries, leading to performance that may fall short of expectations \cite{troumpoukis2020geofedbench}.

Another proposal for including geospatial information in RDF format is stRDF, which is a constraint data model for enhancing RDF's capabilities to represent spatial and temporal data. This proposal also introduces stSPARQL, an extended querying language tailored specifically for navigating stRDF data \cite{koubarakis2010modeling}. An implementation of stSPARQL is Strabon RDF store \cite{kyzirakos2012strabon} which, as mentioned before, also supports the GeoSPARQL query language. SPARQL-st is another data model that defines an upper ontology for modeling spatial and temporal information and extends GeoRSS spatial classes in order to model spatial geometries. It also proposes an extension of SPARQL for complex spatiotemporal queries \cite{perry2011sparql}. A more recent geospatial data model is proposed by stRDFS \cite{zhu2020algebraic} that adds temporal and spatial labels to triples. It also introduces several spatiotemporal classes and defines topological relations and their corresponding spatiotemporal semantics and the spatiotemporal algebraic operations. 

Parallel to the ongoing development of theoretical frameworks for integrating spatial information into RDF stores, numerous projects and research endeavors are dedicated to providing geospatial datasets that adhere to linked data principles. For instance, the LinkedGeoData \cite{stadler2012linkedgeodata} project aims to add a spatial dimension to the Web of Data. It uses the information collected by the Open Street Map project and makes it available as an RDF knowledge base according to the Linked Data principles. A more versatile approach is outlined in \cite{neumaier2019enabling}, where authors introduce a spatial hierarchy to construct knowledge graphs that include geospatial entities and their associated relationships.

In this context, our proposed grid-based approach avoids encoding geometrical features in text-based serialization formats like WKT and GML. Instead, it focuses on translating spatial relationships into RDF entities and spatial hierarchies, enabling direct querying via standard SPARQL. This approach reduces dependence on third-party implementations of RDF stores or SPARQL extensions. A similar grid-based approach is introduced in KnowWhereGraph \cite{janowicz2022know}, which uses the S2 grid system as a base and provides a design pattern for relating features and regions and their interactions via spatial hierarchy. The KnowWhereGraph, instead of traditional linked data approaches that often represent spatial regions as points or polygons, uses grid cells for spatial assignment of entities. Thus, grid cells serve as facilitators for data integration. However, the interconnection between cells and the overlay of cells with road networks, which have been extensively investigated in our work, have not been addressed in KnowWhereGraph. Another type of grid system commonly employed for urban data analysis and smart city applications is the hexagonal grid. For example, in the work presented by \cite{bockling2023wildfire}, a hexagonal spatial grid is utilized to construct a spatiotemporal knowledge graph for the wildfire prediction use case.

\section{Conclusion}
The increasing volume of spatial data underscores the mounting significance of location-based information in today's interconnected world. The ubiquity and criticality of spatial information across various applications have led to an unprecedented surge in the generation of spatial data. Integrating such data into knowledge graphs enriches the representation of real-world entities and fosters deeper insights into their spatial relationships. 

In this research, we introduced grid cells as the foundational element of SKG construction and demonstrated how efficiently the spatial characteristics of real-world entities and their attributes can be encoded within knowledge graphs. The applications presented include a broad spectrum, spanning from the spatial scaling of quantitative indicators to spatial reasoning and inference for overlapping regions. We also introduced a novel methodology for representing street networks in knowledge graphs and the potential of this network for routing and navigation use cases. Then, we demonstrated how this SKG data model is used to address the requirements of a specialized KG for crisis management use cases.

\bibliographystyle{splncs04}
\bibliography{bibliography}

\end{document}